\documentclass[preprint,12pt]{elsarticle}

\usepackage{url}
\usepackage{xcolor}
\usepackage{amsmath}
\usepackage[utf8]{inputenc}
\usepackage{graphicx}
\usepackage{tabularx}
\usepackage[skip=2pt]{caption}
\usepackage{titlesec}
\graphicspath{ {images/} }

\newcommand{\bb}{\textbf{b}}
\newcommand{\br}{\textbf{r}}
\newcommand{\bh}{\textbf{h}}
\newcommand{\bx}{\textbf{x}}
\newcommand{\bW}{\textbf{W}}
\newcommand{\bU}{\textbf{U}}
\newcommand{\bC}{\textbf{C}}

\newcommand{\bff}{\textbf{f}}

\newcommand{\bc}{\textbf{c}}

\newcommand{\bv}{\textbf{v}}

\newcommand{\bi}{\textbf{i}}
\newcommand{\bo}{\textbf{o}}
\newcommand{\bz}{\textbf{z}}
\newcommand{\bX}{\textbf{X}}
\newcommand{\bB}{\textbf{B}}
\newcommand{\bV}{\textbf{V}}
\newcommand{\bLLV}{\textbf{LLV}}
\newcommand{\bLLB}{\textbf{LLB}}

\journal{Image and Vision Computing}

\begin{document}

\begin{frontmatter}



\title{Recurrent Semi-supervised Classification and Constrained Adversarial Generation with Motion Capture Data}


\author[label1,label2]{Félix G. Harvey\corref{cor1}}
\author[label1,label2]{Julien Roy}
\author[label1]{David Kanaa}
\author[label1,label2]{Christopher Pal}

\address[label1]{Polytechnique Montréal,\\ 2900 Édouard-Montpetit blvd,\\ Montreal, H3T 1J4, Canada\\ }
\address[label2]{The Montreal Institute for Learning Algorithms,\\ 2920 Chemin de la Tour,\\ Montreal, H3T 1J4, Canada}

\cortext[cor1]{felix.gingras-harvey@polymtl.ca}

\begin{abstract}
We explore recurrent encoder multi-decoder neural network architectures for semi-supervised sequence classification and reconstruction. We find that the use of multiple reconstruction modules helps models generalize in a classification task when only a small amount of labeled data is available, which is often the case in practice. Such models provide useful high-level representations of motions allowing clustering, searching and faster labeling of new sequences. We also propose a new, realistic partitioning of a well-known, high quality motion-capture dataset for better evaluations. We further explore a novel formulation for future-predicting decoders based on conditional recurrent generative adversarial networks, for which we propose both soft and hard constraints for transition generation derived from desired physical properties of synthesized future movements and desired animation goals. We find that using such constraints allow to stabilize the training of recurrent adversarial architectures for animation generation.
\end{abstract}

\begin{keyword}
Action Recognition, Motion Capture, Semi-supervised Learning, Recurrent Neural Networks, Generative Adversarial Networks, Transition Generation


\end{keyword}

\end{frontmatter}


\section{Introduction}
It is often the case that for a given task only a small amount of labeled data is available compared to a much larger amount of unlabeled data. In these cases, semi-supervised learning may be preferred to supervised learning as it uses all the available data for training, and has good regularization and optimization properties \citep{erhan2010does, bengio2007greedy}. A common technique for semi-supervised learning is to perform training in two phases: unsupervised pre-training, followed by supervised fine tuning \citep{hinton2006fast, bengio2007greedy, erhan2010does, yu2010roles}. The unsupervised pre-training task often consists of training a variant of an auto-encoder (e.g. a denoising auto-encoder) to reconstruct the data. This helps the network bring its initial parameters into a good region of the high-dimensional parameter space before starting to train the model on the supervised task of interest. 

Advances in Recurrent Encoder-Decoder networks have afforded models the ability to perform both supervised learning and unsupervised learning. These architectures are often based on the capacity that recurrent neural networks (RNNs) have to model temporal dependencies in sequential inputs. When handling a sequence, the last hidden state of an RNN can summarize information about the whole sequence, allowing the model to encode input sequences of variable lengths in fixed length vector representations. The separation between the encoder and the decoder networks naturally allows one to easily add, modify or re-purpose decoders for desired tasks. Using multiple decoders forces the encoder to learn rich, multipurpose representations. Additionally, this also allows semi-supervised training in a single phase. 

In this work, we jointly train a classifier model with optional frame-reconstruction, frame-classification, and sequence reconstruction decoders which all affect the sequence representation used by the upper, classification-only layers. Our empirical study shows that adding the unsupervised decoders have a regularizing effect on the supervised sequence classification task when few labeled sequences are available. We demonstrate this improvement on HDM05 \citep{Muller07documentationmocap}, a well known action recognition dataset. We also explore the limits of this method using the more recent NTU-RGB+D dataset \citep{shahroudy2016ntu}. Finally, we perform separate experiments in which a new constrained recurrent adversarial decoder learns to generate future frames conditioned on a similarly encoded past-context representation. We use Long-Short-Term-Memory (LSTM) models to encode and decode sequences, and a simple multilayer perceptron (MLP) to classify them. Our adversarial discriminator is a bi-directional LSTM (BDLSTM), and outputs predictions at each timesteps.

Our main contributions are the following:
We introduce a novel Recurrent Encoder, Multi-Decoder architecture which allows for semi-supervised learning with sequences. We define and execute a set of experiments using more realistic and representative test set partitioning of a widely used public MOCAP dataset, thereby facilitating more informative future evaluations. We show that this updated classification task is still challenging when having an appropriate test set. We show improvements over our implementation of previous state-of-the-art techniques for action recognition on such well defined experiments.
We also present a novel conditional Recurrent Generative Adversarial Network (ReGAN) architecture for predicting future continuous trajectories which integrate multiple physics based constraints as well as desired animation properties. We show that using such data-driven constraints prevents the adversarial learning of the recurrent generator from diverging, greatly improving the generated transitions.

\section{Related Work}
\subsection{MOCAP datasets}
One challenge with the application of deep learning to MOCAP data is the lack of strongly labeled, quality data. For this work, we used two high-quality publicly available MOCAP datasets and a bigger, lower-quality Kinect dataset. The first MOCAP dataset is the HDM05 dataset \cite{Muller07documentationmocap}. It contains $2329$ labeled cuts that are very well suited for action recognition. We use the same 65 classes defined by \citet{cho2013classifying}. The second dataset is the CMU Graphics Lab Motion Capture Database\footnote{http://mocap.cs.cmu.edu/}. This is a significantly bigger MOCAP dataset in terms of number of frames. It contains $2148$ weakly labeled or unlabeled sequences. This dataset can hardly be used for supervised learning as the labeling of sequences, if any, was only made to give high level indications of the actions, and does not seem to have followed any stable conventions throughout the dataset. The work by \citet{zhu2016co}, \citet{ijjina2016classification}, and \citet{barnachon2014ongoing} all use different custom class definitions to obtain quantitative results on CMU for classification. In the present work, we use this dataset for unsupervised learning only. The Kinect dataset is the NTU RGB+D dataset \citep{shahroudy2016ntu} which is to our knowledge the biggest motion dataset containing skeletal motion data. It contains 60 actions performed by 40 different actors, recorded with a Kinect 2 sensor. It consists of more than $56000$ labeled sequences that either contain one or two subjects. Despite the fact that this dataset is approximately 24 times bigger than HDM05 and has a higher actor count, its lower quality and its well defined, realistic partitioning make action recognition in this context a challenging task. Its two evaluation schemes are based on either held-out actors or a held-out view angle. We wish to provide here a similarly well defined evaluation case for HDM05.

\subsection{Recurrent-Encoder-Decoders}
RNNs have proven over the years to be very powerful models for sequential data, such as speech \citep{graves2013speech, sak2014long, graves2013hybrid}, handwriting \citep{graves2008unconstrained}, text \citep{sutskever2011generating, graves2012supervised}, or as in our case, MOCAP \citep{du2015hierarchical, zhu2016co, fragkiadaki2015recurrent, martinez2017human, li2017auto, liu2017skeleton, liu2017global}. We use LSTMs \citep{hochreiter1997long} without in-cell connections (as suggested by \citet{breuel2015benchmarking}) in the models we explore here.
A major advantage and key attribute of RNNs based on Recurrent Encoder-Decoders is their ability to transform variable-length sequences into a fixed-size vector in the encoder, then use one or more decoders to decode this vector for different purposes. Using an RNN as an encoder allows one to obtain this representation of the whole input sequence. \citet{cho2014learning} as well as \citet{sutskever2014sequence} have used this approach for supervised sequence-to-sequence translation, with some differences in the choice of hidden units and in the use of an additional summary vector (and set of weights) in the case of \citet{cho2014learning}. Both approaches need a symbol of end-of-sequence to allow input and target sequences to have different lengths. They are trained to maximize the conditional probability of the target sequence given the input sequence. Our approach is more closely related to the one used by \citet{srivastava2015unsupervised} in which they perform unsupervised learning, by either reconstructing the sequence, predicting the next frames, or both. In our work, an additional decoder is used for classification of whole sequences, and the future generator may use an adversarial loss to improve generated sequences. \citet{mahasseni2016regularizing} also make use of a encoder-decoder LSTM to learn a skeletal motion manifold on small datasets. They then use this manifold to regularize another LSTM performing action recognition on videos (from pixels) by pushing its produced representations to be close to the learned manifold. We focus on skeletal data and combine classification and regularization with a motion manifold in a single phase.

\subsection{Action recognition}
Much of the prior work on MOCAP analysis has been based on hand-crafted features. For example, \citet{chaudhry2013bio} created bio-inspired features based on the findings of \citet{hung2012medial} on the neural encoding of shapes and, using Support Vector Machines (SVMs), have obtained good results on classification of 11 actions from the HDM05. \citet{ijjina2016classification} use some joints distance metrics based on domain knowledge to create features that are then used as inputs to a neural classifier (pre-trained as a stacked auto-encoder). They reach good accuracy for 3 custom classes in the CMU dataset.  Using this prior domain knowledge helps in particular when the dataset is somewhat specialized and may contain actions of a certain type. However, if the goal is to have a generic action classifier that handles at least as many actions as found in HDM05, it might be more appropriate to learn those features with a more complex architecture. \citet{barnachon2014ongoing} use a learned vocabulary of key poses (from K-means) and use distances between histograms of sub-actions in order to classify ongoing actions. They present good accuracy (96.67\%) on a custom subset of 33 actions from HDM05 (where training samples are taken at random). In our case, we wish to perform classification on the 65 HDM05 actions with a realistic test set.

End-to-end neural approaches have also been tried on HDM05 and CMU in which cases discriminating features are learned throughout the training of a neural network. \citet{cho2013classifying} have obtained good movement classification rates on simple sequences (cuts) on the HDM05 dataset using a Multi-Layer Perceptron (MLP) + Auto-Encoder (AE) hybrid. \citet{chen2013classification} tested multiple types of features, using extreme machine learning to classify, again, HDM05 cuts. Results were good in both cases, with accuracies of over 95\% and 92\% with 65 and 40 action classes respectively. Their models were trained at the frame level, and sequence classification was done by majority voting. Other work by \citet{du2015hierarchical} treated the simple sequences' classification problem with the same action classes as \citet{cho2013classifying} with a hierarchical network handling in its first layer parts of the body separately (i.e. torso, arms and legs), and concatenating some of these parts in each layer until the whole body is treated in the last hidden layer. They worked with RNNs to use context information, instead of concatenating features of some previous frames at each timestep. This led to better results, and their classification accuracy on simple sequences reached 96.92\%. Finally, \citet{zhu2016co} have a similar, but less constrained recurrent architecture that is regularized by a weight penalty based on the $l_{2,1}$ norm (\citet{cotter2005sparse}), which encourages parts of the network to focus the most meaningful joints' or features' interactions. They report 97.25\% accuracy on HDM05 for classification of simple sequences, with 65 classes.

Other relevant advances for skeletal motion recognition are applied on different datasets, such as NTU RGB+D, and once again focus on defining new motion data representations \citep{ke2017skeletonnet, ke2017new, vemulapalli2016rolling, zhang2017geometric} in order to inject domain knowledge directly into the inputs. Others propose instead new architectural variants to the neural networks for motion recognition \citep{liu2017skeleton, song2017end, jain2016structural, fragkiadaki2015recurrent, liu2017global} that sometimes induces prior knowledge in the architecture instead. Our own approach could be considered as an architectural modification that aims at reducing the need for domain knowledge by learning better representations for generalization using several decoders. It is in that sense more generic, and could therefore be combined easily with the above approaches or applied to different domains.

\subsection{Generative Adversarial Networks}
GANs \citep{goodfellow2014generative} can be powerful tools to map a random noise distribution to a real data distribution and therefore to generate realistic samples. They are composed of a Generator ($G$) and a discriminator ($D$) that can be both deep neural networks. The goal of $D$ is to tell if a sample comes from the real distribution or if it was generated by $G$ (i.e. it is fake). The generator $G$ learns from the likelihood signal provided by $D$ in order to produce samples closer to real samples. While impressive work has been done with GANs or some of their variants on image generation \citep{reed2016generative, radford2015unsupervised, huang2016stacked}, results on sequential data remains more limited. \citet{ghosh2016contextual} make use of recurrent networks to generate the next plausible image as an answer to a sequence query. In that case, the answer should match the only ground truth answer. \citet{mahasseni2017unsupervised} use an LSTM discriminator to classify reconstructed videos from generated summaries. In our case, we aim at producing a realistic series of positions (which might differ from the true trajectories) that lead to a target pose, conditioned on the compressed representation of the past context, a noise vector, and the target pose itself. During training, our generator and discriminator are not given ground truth frames during their generations/predictions, but always have information about the target pose. For text generation, \citet{yu2016seqgan} use recurrent networks with a policy gradient method to handle discrete outputs. We are interested here in plausible continuous trajectories and thus work with continuous, differentiable, recurrent GANs.

\subsection{Transition Generation}
Approaches have been proposed for motion prediction with well-designed recurrent neural networks \citep{fragkiadaki2015recurrent, martinez2017human, ghosh2017learning, li2017auto} but they don't tackle the transition generation problem where the character need to reach a desired target. Our method also uses an LSTM to generate frames, with added conditioning weights and computations in order to use the target information and the noise vector (in case of adversarial training). Therefore, our proposed constrained adversarial training stabilization procedure could naturally be applied with those motion prediction methods.
Probabilistic models relying on more classical machine learning techniques such as \cite{lehrmann2014efficient} and \cite{wang2008gaussian} have been applied to motion gap-filling, but show limited scalability and are applied to individual types of actions. We apply here a more generic and scalable deep-learning approach based on LSTMs and GANs with additional stabilizing losses, and which has a constant runtime independent from the number of training samples.

\section{Defining a good test set for HDM05}\label{test_sets}
Most previous approaches for classification on HDM05 achieve good classification results when randomly separating the sequences into training, validation and test sets. However, this kind of partitioning is not a fair estimation of the generalization performance of the model, as the network may overfit the action styles of particular actors and perform poorly with new subjects. A more realistic partitioning of HDM05 would therefore be one based on performers, where action recognition accuracy on new subjects can be assessed.

\begin{figure}[h]
\begin{center}
\centerline{\includegraphics[width=.95\textwidth]{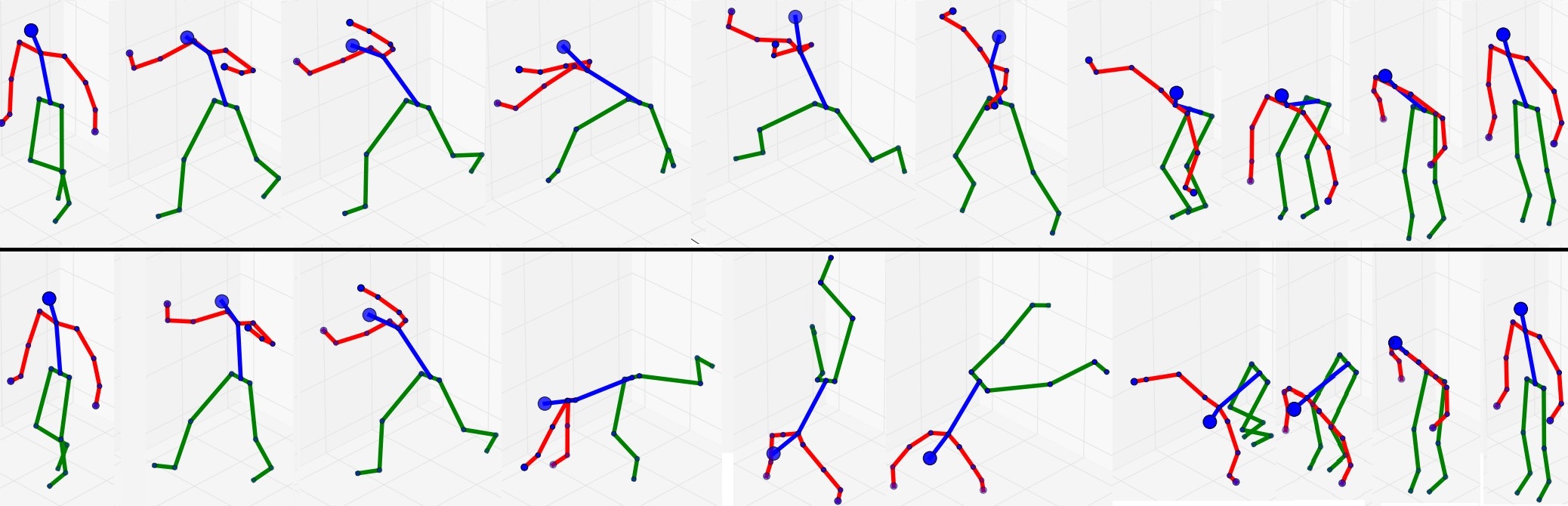}}
  \caption{\small Visual comparison of pre-processing methods for a \textit{cartWheel} movement from HDM05. UP: Same as \citet{cho2013classifying}.   DOWN: our method, that allows for hips not to be parallel to the ground.}
  \label{fig:co}
\end{center}
\end{figure}

\begin{table}[h]
\vspace{.1in}
\caption{Accuracies (Acc.) with different test sets, using techniques from \citet{cho2013classifying}, \citet{du2015hierarchical} and \citet{zhu2016co}. and ours.}
\begin{small} 
\begin{sc}
\begin{center}
\begin{tabular}{llr}
\textbf{Technique}  &\textbf{Test set}  &\textbf{Acc.(\%)}
\\ \hline
Du et al. &Random 10\%, balanced  &92.98\\
Zhu et al. &Random 10\%, balanced &94.53\\
Cho \& Chen &Random 10\%, balanced  &95.61 \\
\vspace{0.05in}
SC (Ours) &Random 10\%, balanced &96.92 \\
\vspace{0.05in}
Cho \& Chen &Random 40\%, balanced  &94.13\\
\vspace{0.05in}
Cho \& Chen &Actors [tr , dg]   &64.36\\
Du et al. &Actors [tr , dg], PP  &70.63\\
Cho \& Chen &Actors [tr , dg], PP   &\textbf{81.64}\\
Zhu et al. &Actors [tr , dg], PP &\textbf{81.64}\\
\end{tabular}
\end{center}
\end{sc}
\end{small}
\label{tab:testComp}
\end{table}
Table \ref{tab:testComp} shows the results of our own implementation of previous state-of-the-art methods \cite{cho2013classifying,du2015hierarchical,zhu2016co} on the HDM05 dataset using our controlled experimental setup. It shows how using held out actors as a test set can hurt the accuracy, and illustrates more clearly that despite the good results of previous methods, action classification on this dataset can still be a challenging task. In this setting, we use actors with initials 'tr' and 'dg' as test subjects, and a random $5\%$ of the training data as a validation set for early stopping and hyper-parameter searches. Since using two out of five actors from HDM05 for testing represents approximately 40\% of the sequences in the dataset, we tested again the method from \citet{cho2013classifying} with a balanced, shuffled partition having the same proportions of sequences in each sets to see if this was the only factor influencing the declining results. Finally, we applied our own pre-processing (PP) of the data with these techniques with our newly defined actor-based partitions to make further comparisons fair. Our preprocessing of the data is explained in Section \ref{res:data} and its effects can be seen in in Figure \ref{fig:co}. As we can see, results using our realistic actors-based partitioning of HDM05 are significantly lower, but our own pre-processing method of the data has a considerable positive effect. Since the techniques of \citet{cho2013classifying} and \citet{zhu2016co} yielded the best results with our actor-based partitions and with our pre-processing method, the baseline test accuracy in the rest of this paper will be of 81.64\% that was reached with those methods. Finally, we also include results from our sequence-classification architecture (SC), which doesn't use any reconstruction with a random $10\%$ balanced test set and using the same preprocessing as \citet{cho2013classifying}.
\section{Our Semi-supervised Classification Models}\label{model}
\begin{figure}[h]
\begin{center}
\centerline{\includegraphics[width=0.95\textwidth]{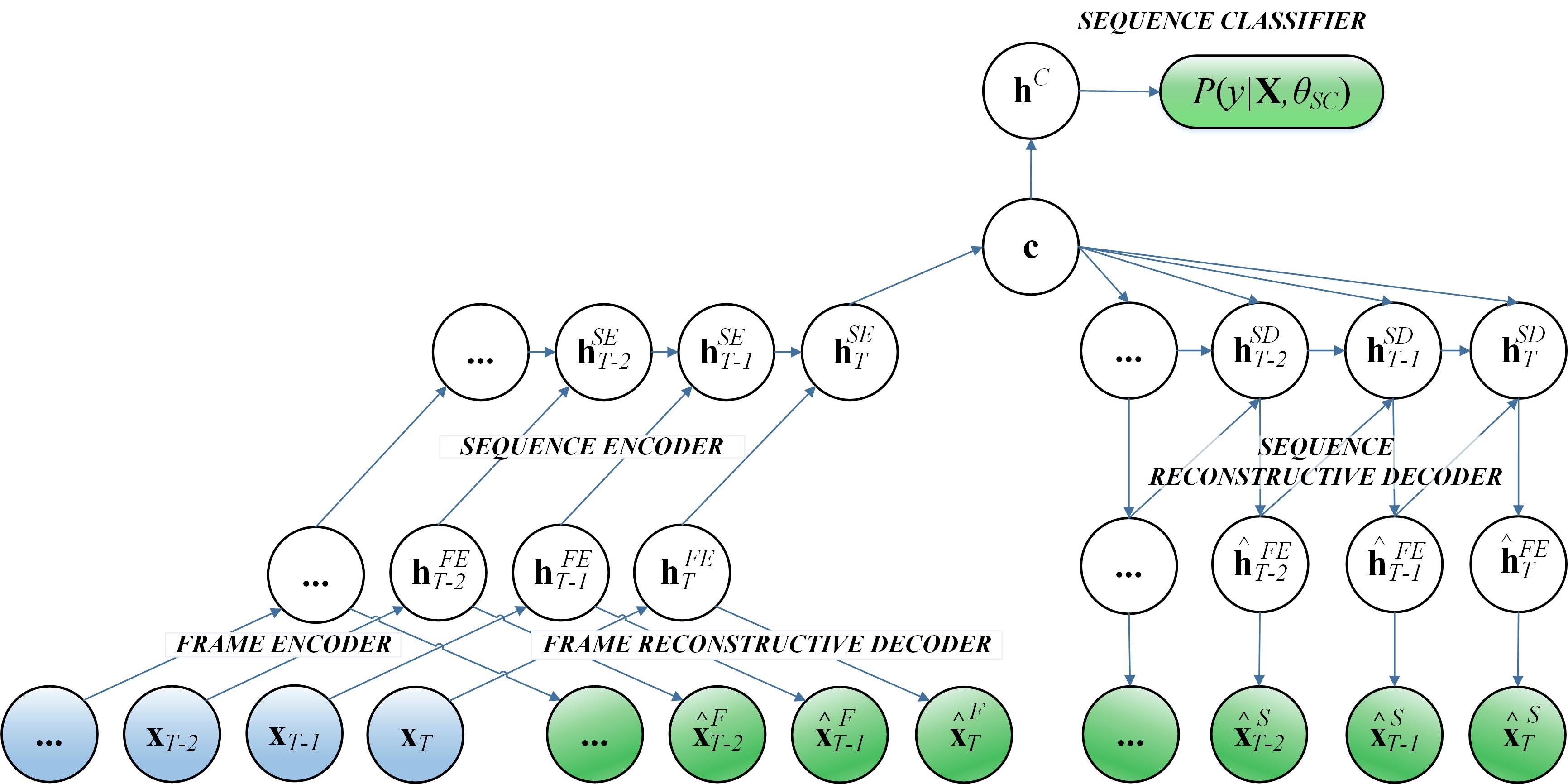}}
\caption{\small The FR-SRC variant of the architecture studied. This network produces 3 types of outputs (in green) with respect to a sequence $\textbf{X}$ = $[\textbf{x}_1, ..., \textbf{x}_T]$ and its parameters $\theta$. The set $\theta_{SC}$ includes all the weights and biases used to compute class probabilities. The hidden states of the frame encoder, sequence encoder and sequence reconstructive decoder are denoted here by $\textbf{h}^{FE}$, $\textbf{h}^{SE}$ and $\textbf{h}^{SD}$ respectively. The sequence representation $\textbf{c}$ is created from the hidden state of the sequence encoder at time $T$, and $\textbf{h}^{c}$ represents the sequence classifier's fully connected layers (the softmax activation is not explicitly shown here).}
\label{fig:complete_faec}
\end{center}
\end{figure}

Figure \ref{fig:complete_faec} shows an overview of the Frame Reconstructive-Sequence Reconstructive Classifier (FR-SRC) variant of the proposed architecture. The model is composed of 5 main components: a per-frame encoder, a per-frame reconstructive decoder, a sequence encoder, a sequence reconstructive decoder, and a sequence classifier. Each decoder along with the classifier produces an output used to calculate a cost. Individual costs have a more direct impact on different modules of the network, and their combination allow to produce evermore meaningful features throughout the model. Note that two additional modules, the future generator and the per-frame classifier, are used in some experiments but are not depctied in Figure~\ref{fig:complete_faec} to avoid cluttering.  The future generator is shown in Section \ref{sec:regan}. The per-frame classifier tries to classify the action based on single frames and takes the per-frame encoding to produce probabilities of actions.
\subsection{Frame reconstruction}
The frame auto-encoder's role is to learn robust per-frame features in an unsupervised manner by trying to reconstruct the clean version ($\bx_t$) of a corrupted frame ($\tilde{\bx}_t$) at time $t$. The reconstructive error ($l_{FR,t}$) we use is the well known mean squared error and we apply it for each frame, before calculating its average over the frames to get $l_{FR}$, where: 
\begin{eqnarray*}
  \bh^{FE}_t &=& u(\tilde{\bx}_t)\\
  \hat{\bx}^F_t &=& g(\bh^{FE}_t)\\
  l_{FR,t} &=& \frac{1}{2}||\hat{\bx}^F_t - \bx_t||^2\\
  l_{FR} &=& \frac{1}{T}\sum_{t=1}^{T}l_{FR,t}
\end{eqnarray*}
Here, $u()$ is the encoding function learned by the bottom feed-forward layers of the per-frame auto-encoder, while $g()$ is the decoding function learned by its upper layers. In further equations, $\textbf{H}^{FE}$ will stand for the sequence of features $[\bh^{FE}_1, ..., \bh^{FE}_T]$ and we will dismiss the corruption sign over $\tilde{\textbf{x}}$ as we will show equations for a test setting, where the frames are not corrupted. All further symbols $\bW$ and $\bb$ without subscript or superscript will represent the weight matrices and bias vectors for the current layer in order to lighten the notation.

\subsection{Frame classification}
The per-frame classifier uses $\bh^{FE}_t$ as an input to yield activations $\textbf{a}_t = \bW(\bh^{FE}_t) + \textbf{b}$ on movement classes for every frame. These activations are then summed over all frames into $\textbf{a}_f = \sum_{t=1}^T \textbf{a}_t\label{b_f}$ and a softmax operation is applied on the result, yielding class probabilities $P(y_k)$ given all the frames $\bx_t$ of the whole sequence $\textbf{X}$, and the parameters of the frame encoder $\theta_{FE}$:
$P(y_k|\textbf{X},\theta_{FE}) = \textbf{s}_{f,k} = e^{\textbf{a}_{f,k}}(\sum_{i=1}^K e^{\textbf{a}_{f,i}})^{-1}$.
%
This is similar to the operation used by Du et al. \cite{du2015hierarchical} to classify sequences based on a sequence of activations but differs in the fact that we do not use outputs from recurrent layers here. We use the negative log-likelihood of the correct class as our frame-classification loss:
\begin{equation*}
    l_{FC}= -log(P(Y=y_k|\textbf{X},\theta_{FE}))
\end{equation*}
The combination of the frame auto-encoder and the frame classifier gives something very similar to Cho \& Chen's \cite{cho2013classifying} approach, except that each frame's input does not contain information about a previous frame. When per-frame reconstruction is not used, the model still encodes frames with $u()$ before outputting probabilities with a softmax.

\subsection{Sequence encoding}
The LSTM encoder's purpose is to encode the whole sequence of learned features into a fixed-length summary vector $\textbf{c}$ that models temporal dependencies, and which can be used for both supervised and unsupervised tasks, 
\begin{equation*}
\textbf{c} = c(\textbf{X}) = \mathrm{tanh}(\bW\bh_T^{SE} + \textbf{b}),
\end{equation*}
where, $c()$ is a fully connected layer parametrized by the weight matrix $\textbf{W}$. It uses the last hidden state of the LSTM encoder $\bh_T^{SE}$ as its input. The encoder itself takes $\textbf{H}^{FE}$ as an input sequence. 

\subsection{Sequence reconstruction}
If the sequence reconstructive decoder is present, it learns to reconstruct the sequence $\textbf{X}$ that was fed to the LSTM encoder. At each timestep, the LSTM decoder uses its previous output along its previous hidden state in order to predict the next frame in an auto-regressive fashion.
With the outputted $\hat{\textbf{X}} = [\hat{\bx}^S_1, ..., \hat{\bx}^S_T]$ from the decoder, and the frame decoding function $g()$, we can calculate our sequence reconstruction error ($l_{SR}$) as a frame-wise MSE loss with the original input sequence:
\begin{gather*}
  \bh_t^{SD}=\mathrm{tanh}(\bW \hat{\bh}^{FE}_{t-1} + \bU\bh^{SD}_{t-1} 
  + \bC\textbf{c} + \textbf{b}) \\
\hat{\bh}^{FE}_{t}=\mathrm{tanh}(\bW\bh_t^{SD} + \textbf{b})\\
\hat{\bx}_t = g(\hat{\bh}^{FE}_t)\label{x_hatt} \\
l_{SR,t} = \frac{1}{2}||\hat{\bx}^S_{t} - \bx_{t}||^2\\
l_{SR} = \frac{1}{T}\sum_{t=1}^{T}l_{SR,t}
\end{gather*}
Here, $\bW$, $\bU$ and $\bC$ are input, recurrent and conditioning weight matrices of the LSTM layer.

\subsection{Sequence classification}
The sequence classifier is a MLP that outputs class probabilities based on the summary vector. This is the main task of interest, the sequence classifier is therefore used in all of our our experiments. We again use the negative log-likelihood as the sequence classification error ($l_{SC}$):
\begin{gather*}
\bh^C = \bW \bc + \textbf{b}, \\ 
\textbf{a}_{seq} = \bW\bh^C + \textbf{b}\\
P(y_k|\textbf{X},\theta_{SC}) = \textbf{s}_{seq,k} = \frac{e^{\textbf{a}_{seq,k}}}{\sum_{i=1}^K e^{\textbf{a}_{seq,i}}}\\
l_{SC} = -log(P(Y=y_k|\textbf{X},\theta_{SC}))
\end{gather*}
\subsection{Total loss}
Using a reconstruction weight $\omega$, we can define different models with different loss functions, enabling some or all of the modules of the architecture. For example, the complete FRC-SRC loss is defined as : 
\begin{gather*}
\ell_{frc-src} = l_{SC} + l_{FC} + \omega \cdot \frac{l_{FR} + l_{SR}}{2}
\end{gather*}
In the generic case, the total loss $\ell$ can be defined as follow:
\begin{gather}
\ell = l_{SC} + i(FC) \cdot l_{FC} + \omega \cdot \frac{i(FR) \cdot l_{FR} + i(SR) \cdot l_{SR}}{i(FR)+i(SR)}
\end{gather}\label{eq_losses}
where the indicator function $i(m)$ is simply equal to $1$ when the module is present and $0$ when it is not. Removing all optional decoders will result in a Sequence Classifier only (SC) network. Adding sequence reconstruction to this model will yield a Sequence Reconstructive Classifier (SRC). Adding instead frame reconstruction to the SC will give a Frame Reconstructive-Sequence Classifier (FR-SC), while adding frame reconstruction to the SRC will yield a Frame Reconstructive SRC (FR-SRC). Finally, adding the last module will result in a Frame Reconstructive Classifier-SRC (FRC-SRC).

\section{Our Constrained Conditional Generation Model} \label{sec:regan}
\begin{figure}[h]
  \centering
  \includegraphics[width=5.0in]{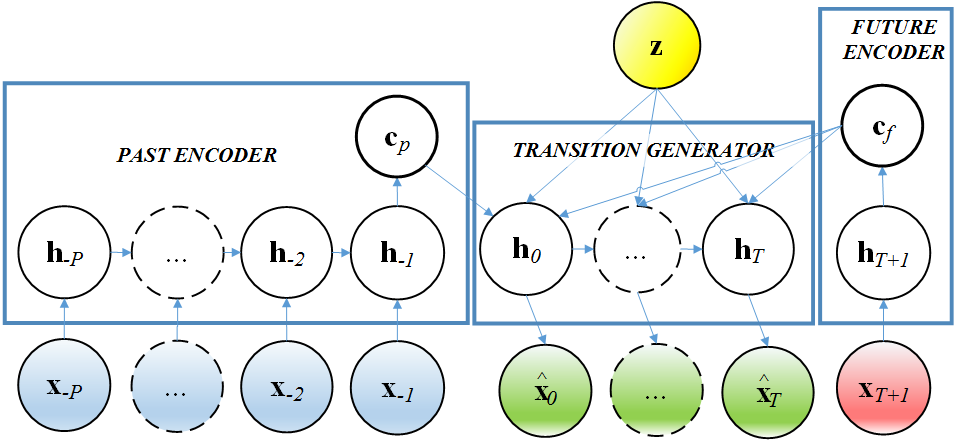}
  \caption{\small Overview of our generative architecture where stacked LSTM layers are not shown. Different time indices are used for this problem.}
  \label{fig:generator}
\end{figure}

As mentioned above, we have also developed a novel constrained conditional recurrent generative adversarial model aimed at creating high quality conditional transition animations. We explored in this context how one could stabilize the training of GANs combined with RNNs, which are both known to be hard to train, by using using physics-based soft constraints forcing the generated clips to respect certain statistics and actual physical constraints of the data. As a tool for generating transitions could be beneficial to animators when desired segments are missing, using GANs for such a task would naturally allow sampling capabilities to such a tool. This motivates our exploration with GANs and why stabilizing their learning could be beneficial. Figure \ref{fig:generator} shows a summary of the generator model. The \emph{past context encoder} has the same structure as the sequence encoder used for classification described above.  We describe the other components below.
%
\vspace{.1in}

\subsection{Future key-pose encoder}
Since the future key-pose is a single vector (single frame), we use here a stack of fully connected layers with no recurrence with the same number of neurons as the past encoder. Similarly to past encoding, we create a future context representation $\bc_f$ that will be used by the generator. With a single layer, $\bc_f = \mathrm{tanh}(\bW\bx_{T+1} + b)$, where $\bx_{T+1}$ is the future key-pose frame at time $t = T+1$.

\subsection{Transition generator}
Our transition generator is a stack of LSTM layers with additional conditioning connections $\bC^{\{i,o,f,c\}}$ that process the conditioning vector $\textbf{r}$, which is a concatenation of the sampled noise vector $\bz$ and the context vector $\bc_{f}$ at each timestep, allowing them to use the target information while generating the transition. The gate and cell values for a single-layer transition generator are computed as follow:

\begin{align*}
\bi_{t} &= \mathrm{sigmoid}(\bW^{(i)} \bx_{t-1} + \bU^{(i)} \bh_{t-1} + \bC^{(i)} \br + \textbf{b}^{(i)})\\
\bo_{t} &= \mathrm{sigmoid}(\bW^{(o)} \bx_{t-1} + \bU^{(o)} \bh_{t-1} + \bC^{(o)} \br + \textbf{b}^{(o)})\\
\bff_{t} &= \mathrm{sigmoid}(\bW^{(f)} \bx_{t-1} + \bU^{(f)} \bh_{t-1} + \bC^{(f)} \br + \textbf{b}^{(f)})\\
\hat{\bc}_{t} &= \bW^{(c)} \bx_{t-1} + \bW^{(c)} \bh_{t-1} + \bC^{(c)} \br + \bb^{(c)}\\
\bc_{t} &= \bff \odot \bc_{t-1} + \bi \odot \mathrm{tanh}(\hat{\bc}_{t})\\
\bh_{t} &= \bo_{t+1} \odot \mathrm{tanh}(\bc_{t})
\end{align*}
where the different $\bW^{\{i,o,f,c\}}$ and $\bU^{\{i,o,f,c\}}$ matrices are the usual input and recurrent connection weights for the different gates and the cell computation, and $\bb^{\{i,o,f,c\}}$ are the bias vectors. The frame output $\bx_t$ can then be computed from $\bh_t$ with the same frame decoding function $\textbf{g}()$ as the one used for sequence reconstruction.

\subsection{Discriminator}
The discriminator consists of a stack of BDLSTM layers that take as input a minibatch of sequences with real and fake transitions and tries to yield higher probabilities for real transitions. The output layer performs a per-frame feed-forward activation $\textbf{a}_t = \bW(\bh^{D}_t) + \textbf{b}$. These activations, for all transition frames, are then summed into a tensor on which the sigmoid classification is done, similarly to what the per-frame classifier does. 

\subsection{Reconstruction objective}
A common way to train a generative architecture is to use a reconstruction loss on the outputted sequences. We can obtain our reconstruction loss with
\begin{equation*}
l_{rec} = \frac{1}{T+1}\sum_{t=0}^{T}\frac{1}{2}||\hat{\bx}_{t} - \bx_t||^2
\end{equation*} 
where $T+1$ is the number of frames in the transition.

\subsection{Adversarial objective}
For the adversarial loss, both our generator $G$ and our discriminator $D$ are needed. Regular adversarial networks \cite{goodfellow2014generative} are designed to be trained by playing a minimax game with
\begin{equation} 
\min_{G} \max_{D} V(D,G) = E_{x \sim p_{data}(x)}[\log D(x)]  
+ E_{z \sim p_{z}(z)}[\log(1-D(G(z)))]. \nonumber
\end{equation}
In our case, however, incoming data to the discriminator is formatted as a sequence $\bX$ of frames containing the past frames $\bX_{past}$, transition frames $\bX_{trans}$ and future frame $\bx_{target}$. Our generator is not only conditioned on the noise vector $\bz$, but also the past context $\bc_p$ and future context $\bc_f$ provided by the past and future encoders. We therefore have the following objective:
\begin{align}
\begin{split}
\min_{G} \max_{D} V(D,G) =&
E_{\bX \sim p_{data}(\bX)}[\log D(\bX_{trans}|\bX_{past}, \bx_{target})]\\
+ &E_{\bz \sim p_{\bz}(\bz)}[\log(1-D(G(\bz,\bc_p,\bc_f)))]
\end{split}
\end{align}\label{gan_obj}

\subsection{Bone length consistency constraint}
Similarly to \cite{holden2015deep}, we apply a bone length consistency constraint in order to preserve rigid bone lengths throughout frames of the generated sequences. In our case, we base our prior knowledge of the bone lengths variations on statistics gathered on the training set. We therefore calculate a vector $\bb^{(m)}$ of mean bone lengths differences between consecutive frames, as well as the vector $\bb^{(v)}$ of variances for all bones differences. This way, we formulate a Gaussian prior (which we expect to be very narrow) on every bone length variations between two frames, which we fit during training using the Log-Likelihood ($\bLLB$) of every bone differences:
\begin{equation}
\bLLB = -\dfrac{1}{2} \log(2\pi\bb^{(v)}) - \dfrac{(\bB - \bb^{(m)})^2}{2\bb^{(v)}}
\end{equation}
where $\bB$ is the matrix of bone length differences on every frame for every bone in the generated sequence, to which the target key-pose is appended. We can therefore retrieve our bone length consistency loss ($L_{bone}$) by averaging all the negative $\bLLB$ :
\begin{equation}
l_{bone} =  \dfrac{1}{N}\dfrac{1}{(T+2)}\sum_{n=0}^{N-1}\sum_{t=0}^{T+1}{ -\bLLB_{n,t}}
\end{equation}\label{lbone}
where $N$ is the number of bones. 

\subsection{Joint velocity constraint}
We also apply a joint velocity constraint based on a mixture of Gaussian priors retrieved from the training set. We perform an EM algorithm to fit two velocity Gaussians for all input dimensions $d$, based on velocities at every frame in the training set. Since bone velocities are very close to 0 on most frames, our mixtures often contain this spike (with a very small variance) and a broader distribution (higher variance). With these mixtures for every joint, we can add a negative log likelihood loss on velocities to constrain bones to have normal velocities, reducing gaps between consecutive frames in the generated transitions. For a given vector $\bv^{(m)}$ of mean velocities per bones and another vector $\bv^{(v)}$ of variances per bones, we get the log-likelihood $\bLLV$:
\begin{equation}
\bLLV = -\dfrac{1}{2} \log(2\pi\bv^{(v)}) - \dfrac{(\bV - \bv^{(m)})^2}{2\bv^{(v)}}
\end{equation}
where $\bV$ is the matrix of velocities of the generated transition to which the target key-pose was appended, for every dimension at every timestep. We can therefore calculate our minimum negative $\bLLV$ for the spike-gaussian and the broader-gaussian velocity statistics and define our loss as:
\begin{equation}
l_{vel} = \dfrac{1}{D}\dfrac{1}{(T+2)}\sum_{d=0}^{D-1}\sum_{t=0}^{T+1}{ \min(-\bLLV^{(spike)}_{d,t}, -\bLLV^{(broad)}_{d,t}) }, \nonumber
\end{equation}\label{lvel}
where $D$ is the number of input dimensions. 

\section{Experimental Setup}\label{setup}
\subsection{Data}\label{res:data}
The data in these experiments comes from three different datasets. The labeled HDM05 dataset and the unlabeled CMU MOCAP dataset are both recorded at 120 frames per second (fps) and contain more than 30 marker positions. In our case, we sub-sample sequences to 30 frames-per-second and use 23 common markers between the two datasets. We work with the C3D file format, which contains series of X, Y, Z positions for each marker, yielding a frame vector of dimension 69. From the NTU dataset, we retrieve the Kinect's skeletal data for each sequences. We use the same approach as \citet{shahroudy2016ntu} to determine the main actor(s) of each sequences. We use the positions of each of the 25 joints for each actor when using this dataset. Since some classes in NTU are two-actor-actions, the standard way of representing the sequences is to concatenate data from the two main actors of each sequences, yielding a data representation with 150 degrees of freedom. In cases where only one actor is present, values for the second actor are kept at 0. We do not sub-sample the NTU sequences as they already have a 30 fps rate.

\subsection{Preprocessing}\label{preprocess}
Our preprocessing of the data for HDM05 and CMU consists mainly of orienting, centering and scaling the point cloud of every frame given by the files. The orientation process is a basis change of all 3D positions so that the actor's hips are always facing the same horizontal direction, while allowing a changing vertical orientation. We then center the hips of the actor at the origin and scale so every marker is always in the interval $[-1,1]$. This can help handling different actors of different sizes. In all experiments on these two datasets, we use an additive Gaussian noise with a standard deviation of 0.05 and mean 0 on markers' positions for training. We use minibatches of size 4 when handling HDM05 only data, 8 when adding CMU data and 32 when using the NTU dataset. On the NTU dataset, our only preprocessing consists of standardizing each joint to have zero mean and unit variance across all the dataset as suggested by \citet{lecun1998efficient}.

\subsection{Network Specifications}
We use a model with a frame encoder that is closely related to the one used by \citet{cho2013classifying}, as it has two hidden layers of $[1024, 512]$ units. Two extra layers of $[1024, 69]$ units are used by the reconstruction decoder with tied weights with the encoder. The LSTM encoder, has 3 hidden layers of $[512,512,256]$ LSTM memory cells. As the output of a single bi-directional recurrent layer can contain, at each timestep, information for the whole sequence, we use bi-directionality only in the first LSTM layer of the sequence encoder. This means that the second layer of the LSTM encoder has an input of size $2*512$ containing past and future information. The $\textbf{c}$ layer, outputting the summary vector is of size $1024$, and the $\textbf{h}^c$ layer is of size $512$. A softmax layer is placed on top of $\textbf{h}^c$ to output action probabilities. Each layer of the LSTM decoder has a number of units equal to the size of the output of its corresponding layer in the encoder. This leads to $[256, 512, 1024]$ memory cells. All non-linear activations used in the network consist of the $\mathrm{tanh}()$ function except for the input, output and forget gates of the memory cells that use sigmoid activations. All reconstructive output layers have linear activations. For feed-forward layers' initializations, their weights are drawn uniformly from $[-\sqrt{1/fanin}, \sqrt{1/fanin}]$, while we use orthonormal initialization for recurrent weight matrices. All biases are initialized at 0, except for LSTM forget gates which are initialized to 1, as proposed by \citet{gers2000learning} and \citet{jozefowicz2015empirical}. 

\subsection{Training Procedure}\label{sec:training}
We use early stopping on the validation set with a tolerance of 20 epochs for HDM05 and 10 epochs for NTU. The learning rate is initialized to 0.04, and is halved when the validation accuracy is not improved for a number of epochs equal to the half of the tolerance. We optimize the network parameters with momentum-augmented stochastic gradient descent with a 0.9 momentum value.

Even though our networks can model sequences of arbitrary lengths, empirical analysis showed that using overlapping sliding windows with a voting strategy to classify a single sequence yielded better results than feeding these complete sequences as a whole to the network. Windows correspond to sub-sequences of a given width, with a constant offset, that are fed to the network once at a time. We then combine the outputs of the network for these windows in order to compute the final classification. More specifically, the network's softmax outputs for all windows of the sequence are summed together and the action class is determined to be the one with the highest added probabilities. This allows for high-activation segments of a full sequence to have a bigger weight in the final classification vote. Before conducting experiments over variations of the classification models, we tested the network using the FR-SRC model on HDM05 data in order to explore different values of reconstruction weights and different sliding window's widths (number of frames we feed to the encoder). We had $\omega \in \{0, 1, 5, 10, 20, 50, 100\}$, where $\omega=0$ means there's no reconstruction, and the window's width w $\in \{20, 30, 90, \infty \}$ where $\infty$ means taking all frames in the sequence. In all other cases, we used an offset of half the width to slide the window. Based on results on our validation set, we found $\omega=50$ and w$=30$ to be most effective. These two hyper-parameters have been fixed to those values for all further experiments. 

The SRC architecture without the classification layers is our Sequence Encoder-Decoder architecture used for generating transitions. It has additional future-conditioning weights in the sequence-decoder to compute each LSTM gate and cell activation. In this case, we modify the targets of the reconstruction to be the future frames of transition. The adversarial learning is standard, where the generator and discriminator alternate updates for optimizing Equation \ref{gan_obj}. When using our soft constraints, the generator also minimizes Equation \ref{lbone} and Equation \ref{lvel} to better shape the generated poses. 

\section{Results}\label{results}
\subsection{Regularizing Classifiers through Reconstruction}
The experiments we conducted here used the small HDM05 dataset only and used our proposed held-out actors as a test set. For these first experiments, we focus on demonstrating the regularizing effects of adding different types of reconstructive modules and losses to the network's composite error function. Table \ref{tab:train_valid_test} shows these effects. Each result is the average classification accuracy for the test set of three different runs with the same architecture.
\begin{table}[h]
\caption{Regularization effects of different models on accuracies\\ with HDM05 data only}
\begin{small}
\begin{sc}
\begin{center}
\begin{tabular}{lrr}
\textbf{Model} &\textbf{Train(\%)}&\textbf{Test(\%)}
\\ \hline
SC        &99.61  &84.08\\
SRC       &99.62  &84.42\\
FR-SC     &99.86  &\textbf{86.14}\\
FR-SRC    &98.83  &85.71\\
FRC-SRC   &98.90  &85.89
\end{tabular}
\end{center}
\end{sc}
\end{small}
\label{tab:train_valid_test}
\end{table}

First, note that our sequence-classifier-only model already outperforms the implemented baselines of 81.64\%. More interestingly, we see that all tested variants with optional modules performed better than our sequence-classifier-only. Specifically, adding only the recurrent reconstruction decoder (SRC) improved only marginally the average performance on the test set, while the biggest improvement came from having per-frame decoders (FR-SC) denoising and improving frame representations for the sequence encoder. It seems however, that in this low-data regime, combining the multiple optional decoders (FR-SRC, FRC-SRC) also help generalization compared to the sequence-classification-only model, but to a lesser extent.

\subsection{Adding Unlabeled CMU Data to Improve HDM05 Classifications}
The following experiments, summarized in Table \ref{tab:models_acc}, compare our results for the movement classification task using HDM05 with and without using additional unlabeled sequences from the CMU dataset. We compare our results with our implementation of the baseline techniques from \citet{cho2013classifying} and \citet{zhu2016co} on the same test set. When adding CMU, we performed the same preprocessing as with HDM05 and, during training, augmented each HDM05 minibatch with an equal number of randomly picked CMU sequences for which no classification cost was computed. As with only labeled sequences, the reconstruction losses were averaged over all sequences in the minibatch.
\vspace{.1in}
\begin{table}[ht]
\caption{Test accuracy of different models}
\begin{small}
\begin{sc}
\begin{center}
\begin{tabular}{llc}
\textbf{Model} &\textbf{Dataset} &\textbf{Test Accuracy (\%)}
\\ \hline
Baseline  &HDM05        &81.64\\
\vspace{.05in}
SC        &HDM05        &84.08\\
SRC       &HDM05        &84.42\\
\vspace{.05in}
SRC       &HDM05+CMU    &84.20\\
FR-SC     &HDM05        &86.14\\
\vspace{.05in}
FR-SC     &HDM05+CMU    &85.71\\
FR-SRC    &HDM05        &85.71\\
\vspace{.05in}
FR-SRC    &HDM05+CMU    &\textbf{87.40}\\
FRC-SRC   &HDM05        &85.89\\
FRC-SRC   &HDM05+CMU    &86.61\\
\end{tabular}
\end{center}
\end{sc}
\end{small}
\label{tab:models_acc}
\end{table}
Since SC does not use any reconstruction, no experiment was done with that model with unlabeled data. The results for HDM05-only experiments from last section are repeated here for easier comparisons. All additional experiments are also averages from three runs in the same setting. In this larger data regime, the use of multiple reconstruction decoders with FR-SRC showed to be the most beneficial addition to the system for generalizing on new subjects. However, when only adding the frame-decoder (FR-SC) or the sequence decoder (SRC), it seems the addition of unlabeled sequences from a different distribution of movements did not help as much as with HDM05 data only. Interestingly, all tested versions of the proposed system showed a similar trend of having CMU data boost performance only when having at least the two reconstructive decoders. We hypothesize that when trying to model the data coming from the different distribution of the CMU classes, too much capacity may be spent on trying to solve the harder job of reconstructing CMU poses when all the weighting of the reconstruction is focused on a single reconstruction task (frame or sequence). However, when combining the reconstruction losses by averaging them (see Equation \ref{eq_losses}), it seems that reduced focus on single objectives helps the network improve the representations by having more generalizable reconstructive features, without wasting its capacity on a single, harder reconstruction task. 
As specified in Section \ref{sec:training}, the search over different $\omega$ values was done only with the FR-SRC model, which could explain why this architectural variant performs best with this weight value when adding unlabeled data.
Also, the frame-based classification module might not be as useful as other modules, which makes sense intuitively. Estimating probabilities of an action based on a single frame without context might be a task overly complex - or simply impossible - for the network. Therefore, the per-frame encoding layers might try to reduce the very high loss on frame-based classification by (often unsuccessfully) producing discriminative features at the expense of higher other losses, resulting in less useful features to sent the LSTM encoder.

\begin{figure}[t]
\smallskip
\begin{center}
\centerline{\includegraphics[width=0.9\textwidth]{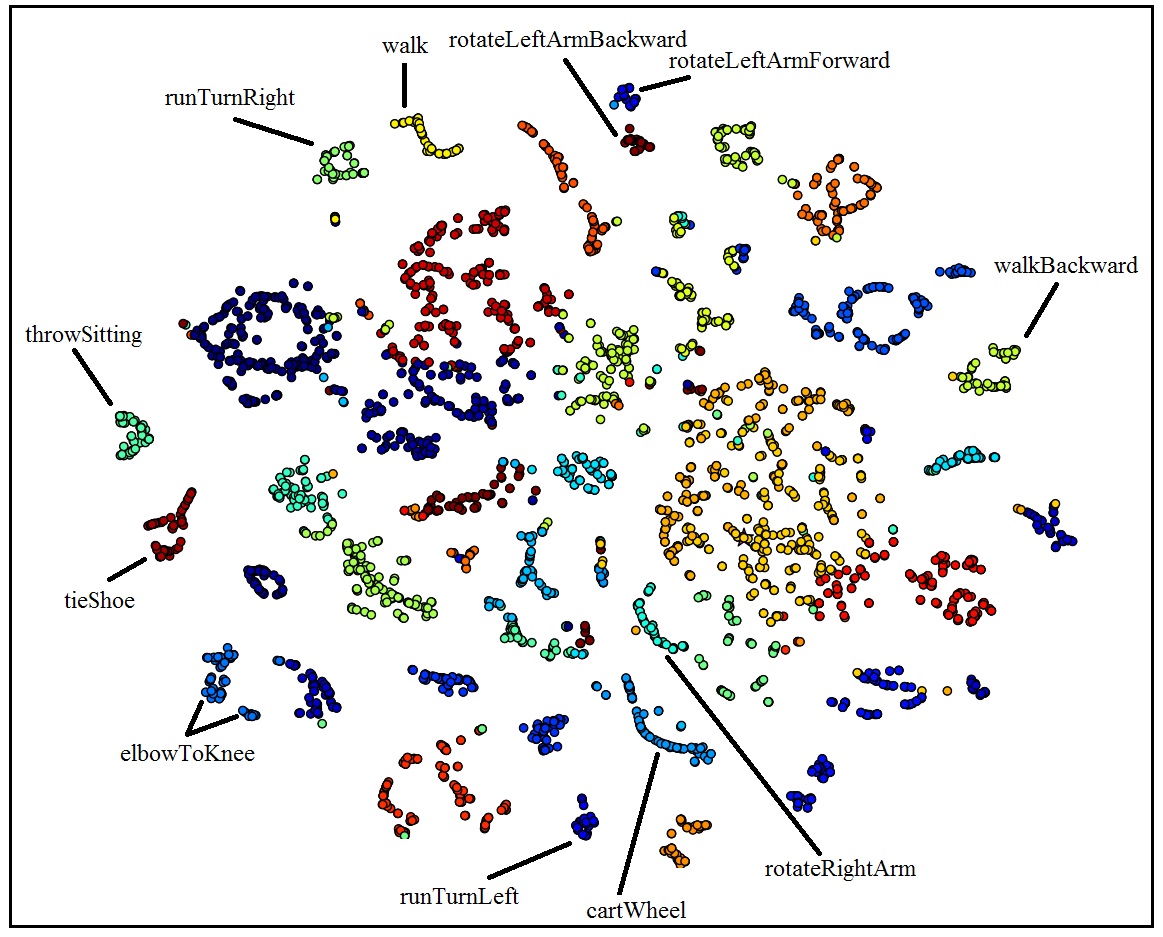}}
  \caption{2D visualization of clusters found by the FR-SRC network with some hand-made annotations after inspection of sub-sequences inside clusters.}
  \label{fig:clusters}
\end{center}
\vspace{-.15in}
\end{figure}

\subsection{Clustering HDM05} Using the FR-SRC network that yielded the best results on HDM05 classification, we performed clustering on the summary vectors it produced for the test set, unseen during training. We used a Gaussian Mixture Model (GMM) initialized with the K-means++ algorithm, where K was found by using 10\% of the set as a validation set to find the best likelihood. This system found 30 clusters that we can visualize in Figure \ref{fig:clusters}. Note that feature vectors have 1024 dimensions and clusters were found in that space, while we used the t-SNE algorithm \citep{maaten2008visualizing} to create a 2D visualization. Some clusters were annotated after manual inspection to give an idea of what movements the network clustered. We can see that such a trained network could help accelerate labeling MOCAP sequences of movements since sequences in the most well defined clusters could be labeled in batch. Manual annotation seems to suggest that HDM05 actions have a considerable impact of the clustered actions, since almost all clusters could be associated with one or two HDM05 labels.

\subsection{Exploring the Higher Data Regime}

To study the effect of our approach in a higher data regime, we evaluate our models on NTU RGB+D \citep{shahroudy2016ntu}, a more recent and much larger dataset. It contains 56,880 action sequences (compared to 2329 for HDM05), recorded on 40 different actors and from 3 different camera views. Our results on this dataset are presented in table~\ref{tab:NTU_results}. 

\begin{table}[ht]
\caption{Performance in terms of accuracy on the NTU RGB+D dataset}
\begin{small}
\begin{sc}
\begin{center}
\begin{tabular}{p{7cm} ll}
\textbf{Model} &\textbf{CS} &\textbf{CV} \\
\hline
Squeletal Quads \citep{evangelidis2014skeletal}  &38.6       &41.4\\
Lie Group \citep{vemulapalli2014human}    &50.1       &52.8\\
FTP Dynamic Skeletons \citep{hu2015jointly}     &60.2       &65.2\\
HBDRNN \citep{du2015hierarchical}  &59.1       &64.0\\
Deep RNN \citep{shahroudy2016ntu}               &56.3       &64.1\\
Deep LSTM \citep{shahroudy2016ntu}            &60.7       &67.3\\
Part-aware LSTM \citep{shahroudy2016ntu}  &62.9       &70.3\\
ST-LSTM \citep{liu2017skeleton} &69.2       &77.7\\
STA-LSTM \citep{song2017end} &73.4       &81.2\\
CNN-MTLN \citep{ke2017new}  &\textbf{79.6}       &\textbf{84.8}\\
\hline
SC    &62.5    &65.88 \\
FR-SC   &63.4        &65.60 \\
\end{tabular}
\end{center}
\end{sc}
\end{small}
\label{tab:NTU_results}
\end{table}

Our models, trained with the same hyper-parameters as with the other datasets, achieve similar results (slightly better on cross-subject evaluation and slightly worse on cross-views) to those of the Deep LSTM \citep{shahroudy2016ntu} originally proposed as a baseline for NTU. Table~\ref{tab:NTU_results} shows that even though adding frame reconstruction seem to improve our results on cross-subject evaluation, it does the opposite on cross-views, supporting the idea that its effect is mitigated in higher data regimes. Also, it hints at the fact that extra regularization strategies, such as ours, can help to generalize to new actors, but are not required when the test-set is composed of almost identical sequences to the ones found in the training set, as it is the case with the preprocessed NTU sequences in the cross-views evaluation. In other words, additional unsupervised objectives clearly show to be beneficial on HDM05 and CMU datasets, as supported by Table~\ref{tab:models_acc}, but do not provide the same benefit on much larger datasets. We hypothesize that such behavior could emerge from the fact that our multi-decoder approach effectively improves generalization on smaller datasets by acting as a regularization strategy, as supported by Table~\ref{tab:train_valid_test}, and that even though action classification on NTU is still a very challenging task, the large number of sequences and the impressive variety of actors included in the training set reduce the need for that kind of regularizer.

\subsection{Constrained Adversarial Generation}
\begin{figure*}[ht]
\begin{center}
\includegraphics[width=1.0\textwidth]{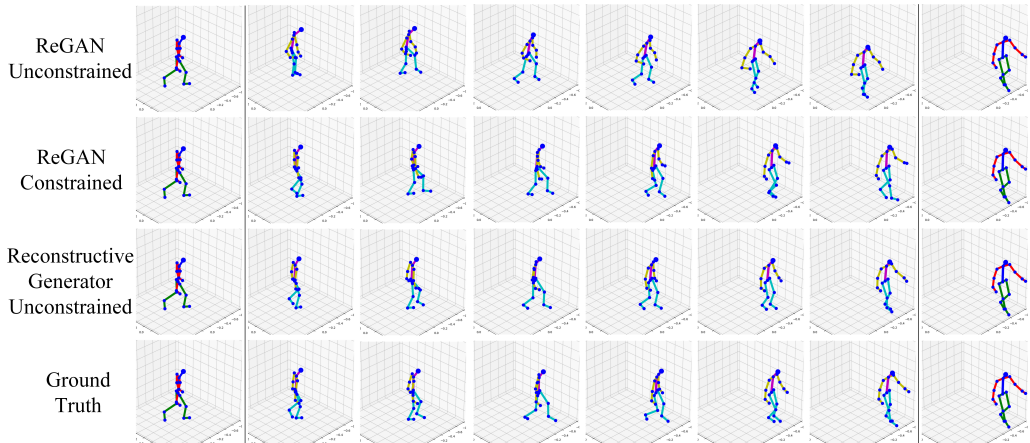}
  \caption{\small Comparison of different generated \textit{Walk-Turn} transitions with the ground truth. Differences between the generated transitions are easier to observe in the supplementary video.}
  \label{fig:generations}
\end{center}
\end{figure*}
\begin{figure*}[ht]
\begin{center}
\includegraphics[width=0.92\textwidth]{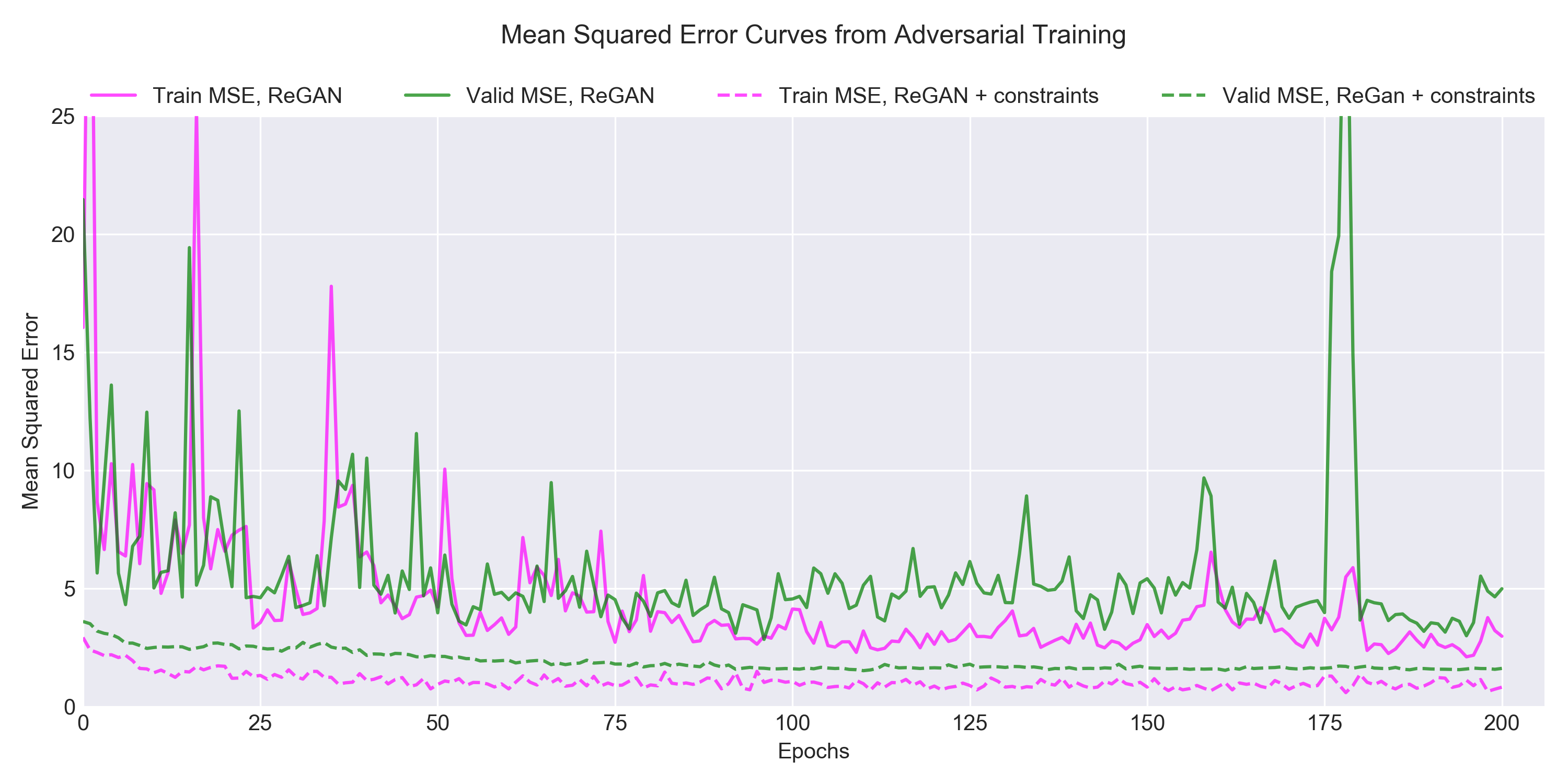}
  \caption{\small Mean squared error (MSE) curves comparison from adversarial training, with and without constraints. Our added soft constraints stabilize and improve performance throughout training.}
  \label{fig:adv_mse}
\end{center}
\end{figure*}

We apply our proposed conditional future generator on the combination of the HDM05 and CMU datasets. In this setup, we use $15$ seed frames to produce the past context vector $\bc_p$ and generate $30$ transition frames that lead to a target frame encoded in $\bc_f$. Figure \ref{fig:generations} shows a sample output with different losses. The improvement due to the addition of physics-based soft constraints is easily noticeable as without them the generator never learns to produce smooth transitions and often exhibits physically implausible artifacts, such as stretched bones or jumps between frames. Since differences between samples from the reconstructive and constrained adversarial generators (for example, small reductions of foot sliding) can be difficult to identify on still frames, the reader is encouraged to watch the videos in the supplementary materials. The effects of adding our data-driven soft constraints during training are also depicted in Figure \ref{fig:adv_mse}, where MSE curves are shown for an adversarial training procedure.
Even though we do not directly optimize to minimize the MSE, Figures \ref{fig:generations} and \ref{fig:adv_mse} together show that over a certain MSE threshold, the generated transitions are neither realistic nor smooth. We further quantify the effects of both our proposed constraints ($l_{bone}$ and $l_{vel}$) in an ablation study reported in Table \ref{tab:ablation}.
\begin{table}[h]
\caption{Ablation study for the soft-constraints with adversarial training.}
\begin{small}
\begin{sc}
\begin{center}
\begin{tabular}{lrr}
\textbf{Model} &\textbf{MSE}
\\ \hline
ReGAN                               &0.355\\
ReGAN + $l_{bone}$                  &0.163\\
ReGAN + $l_{vel}$                   &0.137\\
ReGAN + $l_{vel}$ + $l_{bone}$    &\textbf{0.116}\\
\end{tabular}
\end{center}
\end{sc}
\end{small}
\label{tab:ablation}
\end{table}
%
%
%
In summary, the main contribution of these exploratory experiments is the significant improvement of stability for recurrent adversarial learning brought by the soft constraints based on training-data statistics that do not impose a deterministic and  unique path given a past and a future context. This can be used in other settings as it is a good way to enforce realism without diminishing the ability of GANs to produce realistic samples untied to a single path. It also significantly makes the training of the combination of LSTM and GANs much more stable.

\section{Conclusion}\label{sec:conclusion}
Recurrent Encoder-Decoder architectures with multiple decoders provide an attractive framework for semi-supervised, multi-purpose representation learning. Our experiments show that the explored architectures outperform our implementations of the state-of-the-art for HDM05 movement classification methods with a realistic actor-based partition of data. We hope this evaluation setup can serve as a benchmark partition for further HDM05 experiments, which is still a challenging dataset due to its high number of action classes and low actor count. Our results also indicate that the inclusion of reconstructive decoders can have a regularizing effect on learning and allow the use of unlabeled data in order to improve generalization. Additionally, we have seen that such networks are well suited for clustering as learned representations compress both reconstructive and discriminative information about sequences. Clusters therefore tend to correspond to actions that resemble labels and could therefore accelerate further labeling.

Our experiments on NTU RGB+D dataset, which contains both many more sequences and more actors, show the limited beneficial aspects of our unsupervised-decoders on larger datasets. This indicates that our approach is most useful when working with a limited amount of labeled data, which is still quite common for real-life applications.

Finally we have also explored a novel constrained recurrent adversarial animation transition generator which can produce plausible continuous skeletal trajectories. Both LSTMs and GANs can be hard to train, but we observed clear benefits from the addition of soft constraints with adversarial training and believe this points to promising directions for realistic animation synthesis and continuous, variable length trajectory generation. The idea of defining data-driven soft constraint could be applied to other temporal domains where GANs are especially hard to train.

\section*{Acknowledgements}\label{sec:ack}
We thank Ubisoft and the Natural Sciences and Engineering Research Council of Canada for support under the Collaborative Research and Development program. We also want to thank the authors of the Theano framework \citep{team2016theano}.




\end{document}